\documentclass[a4paper]{article}

\usepackage{algorithm}
\usepackage[noend]{algpseudocode}
\usepackage[english]{babel}
\usepackage[utf8x]{inputenc}
\usepackage[T1]{fontenc}
\usepackage{mathrsfs}
\usepackage{amsfonts}
\usepackage{float}
\usepackage{mwe}
\usepackage{natbib}
\usepackage{graphicx}
\usepackage{subfig}
\usepackage{mwe}
\usepackage{bbm}
\usepackage[a4paper,top=3cm,bottom=2cm,left=3cm,right=3cm,marginparwidth=1.75cm]{geometry}

\usepackage{amsmath}
\usepackage{graphicx}
\usepackage[colorinlistoftodos]{todonotes}
\usepackage[colorlinks=true, allcolors=blue]{hyperref}
\usepackage{color}

\usepackage{amsthm}
 
 \makeatletter
\def\BState{\State\hskip-\ALG@thistlm}
\makeatother
\theoremstyle{definition}
\DeclareMathOperator*{\argmin}{argmin}
\DeclareMathOperator*{\argmax}{argmax}
\DeclareMathOperator*{\mse}{mse}

\DeclareMathOperator{\E}{\mathbb{E}}
\DeclareMathOperator{\R}{\mathbb{R}}
\DeclareMathOperator*{\error}{error}

\title{Bayesian Optimization Using Monotonicity Information and Its Application in Machine Learning Hyperparameter Tuning}
\begin{document}
\author{Wenyi Wang \\ Computer Science Department, University of British Columbia\\ \href{mailto:wenyw@cs.ubc.ca}{wenyw@cs.ubc.ca} 
   \and William J. Welch \\ Statistics Deparment, University of British Columbia\\ \href{mailto:will@stat.ubc.ca}{will@stat.ubc.ca} }
\maketitle

\begin{abstract}
We propose an algorithm for a family of optimization problems where the objective can be decomposed as a sum of functions with monotonicity properties.  The motivating problem is optimization of hyperparameters of machine learning algorithms, where we argue that the objective, validation error, can be decomposed as monotonic functions of the hyperparameters.  Our proposed algorithm adapts Bayesian optimization methods to incorporate the monotonicity constraints.  We illustrate the advantages of exploiting monotonicity using illustrative examples and demonstrate the improvements in optimization efficiency for some machine learning hyperparameter tuning applications.
\end{abstract}

\section{Introduction}

Bayesian optimization has been successfully applied to many global optimization problems \citep{jones1998efficient, martinez2007active, hutter2011sequential, snoek2012practical}. Typically, it makes few assumptions about the objective function, treating it as a black box.  When prior knowledge is available, however, it might be possible to improve the efficiency of the optimization search. In particular function monotonicity has been successfully exploited to improve statistical modeling, \citep[e.g.,][]{golchi2015monotone} for analysis of computer experiments. The methods proposed here employ such monotonicity information for problems motivated by machine learning (ML), where the performance of an ML algorithm model is complex with respect its hyperparameters, which have to be tuned. We propose a sequential method that adapts the Bayesian optimization framework for an objective function that can be decomposed into a sum of functions with monotonicity constraints and exploit that structure. We analyze the method's applicability to ML hyperparameter problems and provide positive experimental results.

Our algorithm incorporates monotonicity information in the Gaussian process (GP) model underlying Bayesian optimization. Bayesian optimization sequentially adds new objective function evaluations based on a probabilistic emulation of the objective trained using all current evaluations.  At each iteration the emulator guides the choice of the next evaluation in a way that balances global and local search. A common choice for the statistical emulator is GP regression. In our work, instead of modeling the objective function directly by a GP, we decompose the function into components emulated by approximately monotonic GPs.

The rest of this paper is organized in the following way. In section \ref{sec:reviews} we provide a brief overview of Bayesian optimization and GPs with monotonicity constraints, and discuss a related work on shape constrained hyperparameter optimization. Section \ref{sec:formAndAlg} illustrates the algorithm's basic ideas of objective decomposition and monotonicity with a simple example. In section \ref{sec:app} we argue that these ideas can be applied to ML hyperparameter tuning, and compare the proposed algorithm with a standard Bayesian optimization approach. Finally, in section \ref{sec:last} we make some concluding remarks and discuss future work.

\section{Related Works}
\label{sec:reviews}

\subsection{Bayesian Optimization}
\label{sec:BO}
Bayesian optimization is a sequential model based algorithm. At each iteration it builds a probabilistic model of the objective function based on previous evaluations, and chooses the location of the next function evaluation based on the model. Unlike local search strategies, it tends to use more (or all for black-box problems) available information to determine the next function evaluation. This makes Bayesian optimization particular useful when few objective function evaluations are practical, without knowledge of the properties (e.g. convexity) of the objective function.

Algorithm \ref{alg:BO} describes a typical procedure for  Bayesian optimization. The algorithm takes the region of interest $S \subset \R^d$, the objective function $f: S \to \R$, a set $I$ of vectors from $S$, a model fitting function $M$, and an acquisition function $A$ as input, and outputs a list of objective function evaluations. First it initializes objective function evaluation set $D$ at $I$. At each iterative step it builds a model $h$ based on $D$, finds the next evaluation point $X_i$ that optimizes $A(\cdot, h)$, and append  $(X_i, f(X_i))$ to $D$. Common choices of $M$ include GP regression and random forest. Expected improvement, probability of improvement, and upper confidence bound are typically used for the acquisition function. For a comprehensive overview see \citet{brochu2010tutorial}. In this paper we focus on GP regression for fitting probabilistic models, and adapt expected improvement acquisition function. 

\begin{algorithm}
\caption{Bayesian Optimization}\label{alg:BO}
\begin{algorithmic}[1]
\State \textbf{Input:} objective function $f$, region of interests $S$, model fitting function $M$ and acquisition function $A$
\State \text{Output:} D: a list of objective function evaluations
\State Initialize the training data $D = \{(X,f(X)) : X \in I\}$ 
\State \textbf{repeat}
\State \quad Fit a probabilistic model $h = M(D)$
\State \quad Determine the next evaluation point at $X = \argmax_{x\in S} A(x,h)$
\State \quad Append $D=D \cup {(X, f(X))}$
\State \textbf{until} maximum number of iterations reached or other stopping criteria satisfied 
\State \textbf{return} D
\end{algorithmic}
\end{algorithm}

\subsection{Gaussian Processes with Monotonicity Constraints}

As discussed in section \ref{sec:BO}, GP regression builds a probabilistic model of a function based on its point evaluations. It is a robust model that can be applied to a large family of functions including non smooth, non continuous, or discrete functions. In particular, it assumes the target function is sampled from a GP, fits parameters that are used to describe the GP, and applies conditional distribution for inference. For more details of GP regression see \citet{rasmussen2006gaussian}.

When priori knowledge of the target function is available, performance of the model can be improved. In this paper, we study functions that can be decomposed as a sum of monotonic functions, and model the decomposed functions separately. There are various approaches to GP modeling subject to (approximate) constraints \citep{riihimaki2010gaussian, wang2016estimating, lin2014bayesian}. We employ \citet{riihimaki2010gaussian}'s implementation because it is readily accessible. Alternative methods will be discussed in section \ref{sec:last}.

\citet{riihimaki2010gaussian} enforce monotonicity by introducing virtual derivative observations. They introduce a set $V$ of virtual points $x$, and assume  positive (or negative) derivatives are observed at these points. For definiteness, take a target function that increases with respect to $x_j$.
Conditional on the true derivative at point $V_i$ in the  $x_j$ direction,
the prior probability of the event $M_{i,j}$ that the observed (with noise) derivative is  positive equals  
\begin{align*}
P(M_{i,j} = \text{true} | \frac{\partial f}{\partial X_j}(V_i) = z) = \Phi \big( \frac{z}{\nu} \big),
\end{align*}
where $\Phi$ is the cumulative distribution function of the standard normal distribution and  $\nu$ is a constant scalar. For the experiments of this paper, we set $\nu = 0.1$. For reasonably smooth GPs, to incorporate monotonicity information, it is sufficient to add positive (or negative) derivative observations at a modest number of virtual points.

With this prior probability, and knowing the fact that the derivatives of a GP is also a GP, a conditional GP can be approximated by expectation propagation. Therefore parameters can be learned using maximum a posterior estimate, and inference are performed as conditional density estimation.

\subsection{Bayesian Optimization with Shape Constraints}
Another work has been studied that enforce shape constraint on the objective function for Bayesian optimization \citep{jauch2016bayesian}. Instead of decomposing the objective function into sum of monotonic functions, \citet{jauch2016bayesian} assuming constrains include monotonicity, convexity and quasiconvexity directly to the objective function. We believe these constrains are inappropriate for the following reasons. For monotonic objective functions, the region of interests can be reduced, and monotonicity is useless in the reduced region of interests. For convex objective functions, there exists more sufficient algorithms for optimization. For quasiconvex functions, the paper constrains the objective function by enforce  different monotonicities on each side of critical points. However it does not illustrate how to find the critical points, and given critical points the optimization problem is reduced to much easier problem. For all the three constraints, the restricted objective function are too limited. To the best of our knowledge, for ML hyperparameter tuning problems, such properties of the objective function does not exist. On experimental side, their results demonstrate that the best predicated mean and variance of the objective function are better than those predicted by standard method. However we do not see a result on evaluated objective function value.

\section{Problem Formulation and Proposed Algorithm}
\label{sec:formAndAlg}
In this section we define the problem of interests and describe our algorithm rigorously.  An example of applying our algorithm on an artificially designed problem is provided.
\subsection{Problem Definition and Algorithm Description}
Our focus is an objective function $f$ that can be written as a sum of functions, 
\[
f(x) = \sum_i f_i(x), 
\]
with monotonicity constraints 
\begin{align*}
& f_i(x_1, \dots, x_{j-1}, x_j, x_{j+1}, \dots, x_n) \\
\leq & f_i(x_1, \dots, x_{j-1}, x_j', x_{j+1}, \dots, x_n)
\end{align*}
or
\begin{align*}
& f_i(x_1, \dots, x_{j-1}, x_j, x_{j+1}, \dots, x_n) \\
\geq &f_i(x_1, \dots, x_{j-1}, x_j', x_{j+1}, \dots, x_n)
\end{align*}
for some $i,j$ and all feasible $x_j < x_j'$ and $(x_1, \dots, x_{j-1}, x_{j+1}, \dots, x_n)$.
Whether the function $f_i(x)$ is monotonically non-decreasing or non-increasing with respect to $x_j$ must be known.
Our algorithm needs to observe each function $f_i(x)$ at chosen $x$ values in the domain.
We assume such observations are contaminated by Gaussian noises. 

Our algorithm works in the same way as the Bayesian optimization framework described in section \ref{sec:BO}, with adaptation of the modeling step. Instead of modeling $f(x)$ directly, we model each $f_i(x)$ separately using independent monotonicity-constrained GP regressions. 
The predictive mean of $f(x)$ at any trial $x$ is then the sum of the predictive means of the $f_i(x)$, and because of the assumed independence of the component GPs, the predictive variance of the sum is similarly the sum of the variances.
Once the predictive mean and variance of $f(x)$ has been computed in this way,
the expected improvement of a new evaluation at $x$ can be computed in the standard way.

For all the experiments presented, each component GP is assumed to have zero mean, constant variance, and isotropic squared-exponential correlation function, as well as a monotonicity constraint if known.  The GP also assumes a Gaussian-noised oracle. The search starts from four initial evaluations since there are three parameters to be estimated for each GP regression: the variance of the correlated part of the GP, the scale parameter of the squared exponential correlation function, and the variance of the independent noise.

\subsection{Illustrative Example}
\label{sec:NR}
A simple example will demonstrate the proposed algorithm and show that it can be more efficient than a standard Bayesian optimization approach. To examine the roles of function decomposition and monotonicity, we also compare with a modification of our algorithm without monotonicity modeling. Thus, the methods compared are referred to as (1) standard, (2) decomposition and monotonicity (the proposed algorithm), and (3) decomposition without monotonicity.

\begin{figure}[h]
\includegraphics[width=1.0\columnwidth]{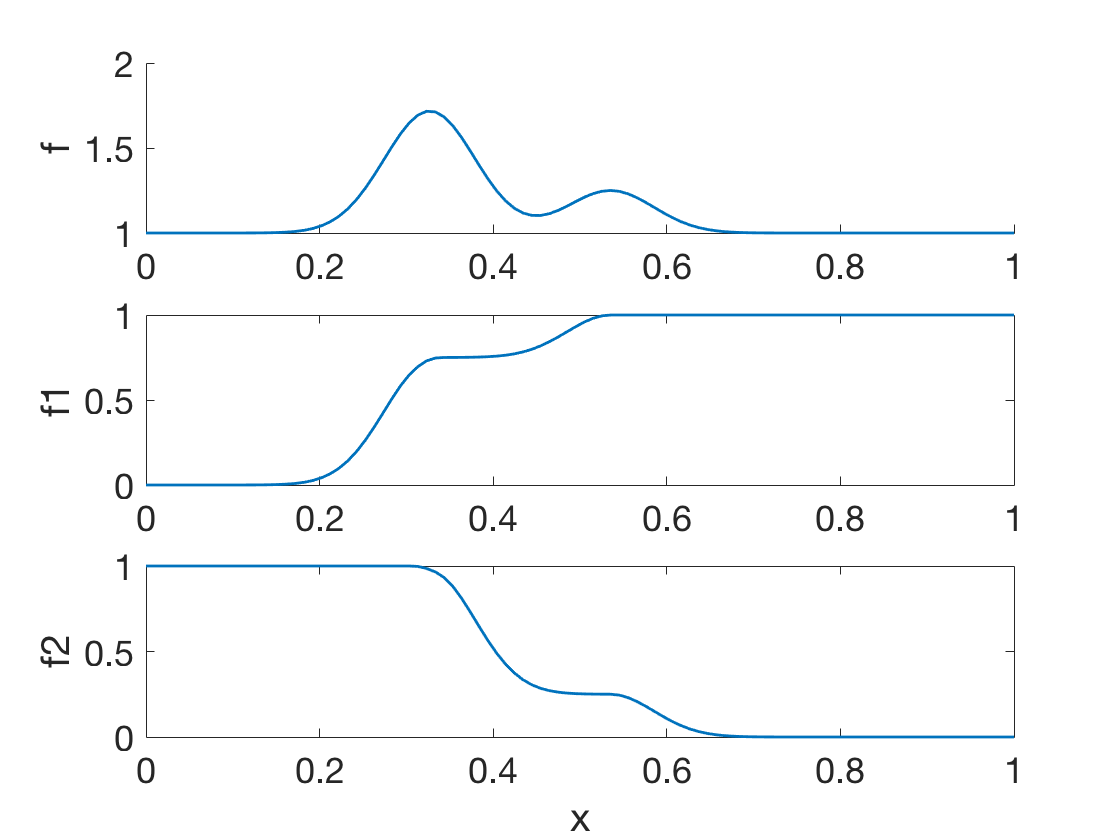}
\caption{Illustrative objective function $f(x)$ and its decomposition into $f_1(x) + f_2(x)$. }
\label{fig:1d_obj}
\end{figure}

The illustrative example is to maximize
\[
\frac{1}{|{\cal X}|\phi(0,\sigma)}\sum_{x' \in {\cal X}} \phi(x-x',\sigma) \quad (0 \leq x  \leq 1),
\]
where $\phi(y,\sigma) = \frac{1}{\sigma \sqrt{2\pi}} e^{-y^2 / (2 \sigma^2)}$ is a basis function with the shape of a probability density function of the normal distribution with mean zero and standard deviation $\sigma = 0.05$. The basis functions are centered at ${\cal X} = \{0.5351, 0.3412, 0.3061, 0.3325\}$. Without changing the problem by adding a constant, consider the objective function 
\[
f(x) = 1+\frac{1}{|{\cal X}|\phi(0,\sigma)}\sum_{x' \in {\cal X}} \phi(x-x',\sigma) \quad (0 \leq x  \leq 1).
\]
One can show that  $f(x)$ is the sum of two monotonic functions $f_1$ and $f_2$, where
\begin{align*}
f_1(x) &= \frac{1}{|X|\phi(0,\sigma)}\sum_{x' \in X} \phi(\max(x-x',0),\sigma), \quad \text{and} \\
f_2(x) &= \frac{1}{|X|\phi(0,\sigma)}\sum_{x' \in X} \phi(\min(x-x',0),\sigma).
\end{align*}
The objective function and its decomposition are shown in Figure \ref{fig:1d_obj}.

\begin{figure}[h]
\centering
\subfloat[Standard]
{
	\includegraphics[width=.49\linewidth]{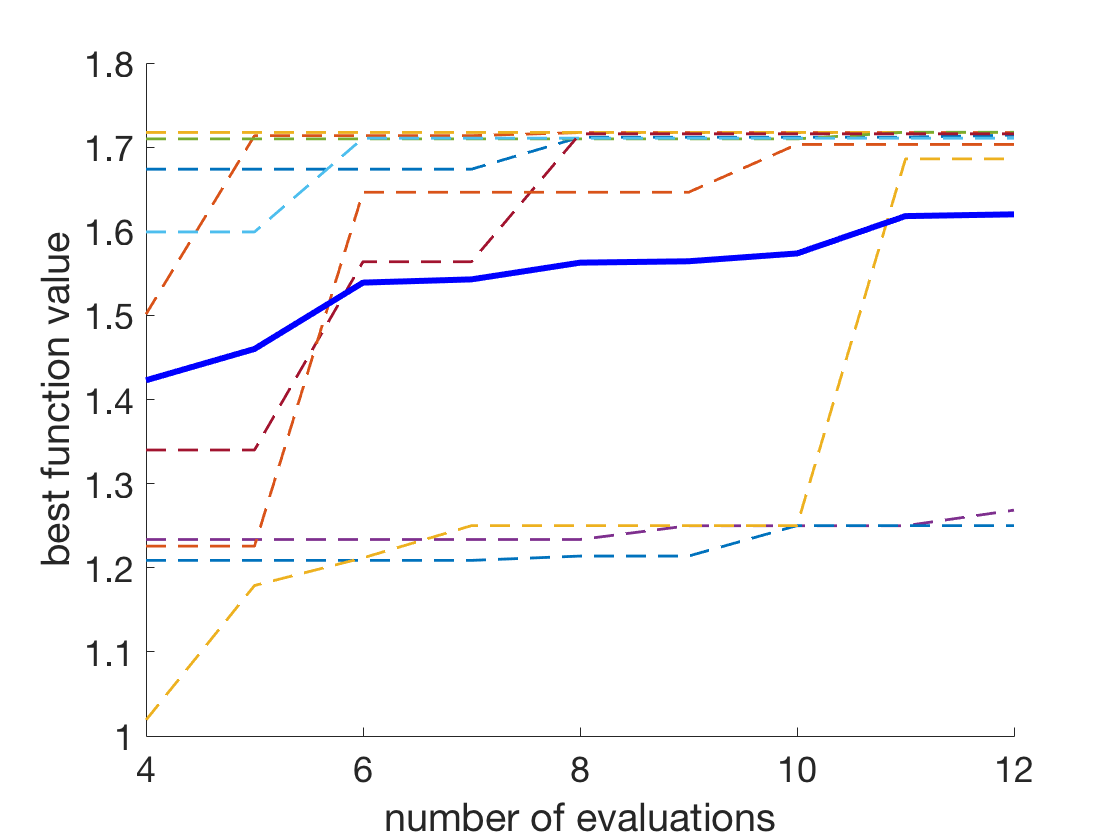}
	\label{fig:1d_standard}
}
\subfloat[Decomposition and monotonicity]
{
	\includegraphics[width=.49\linewidth]{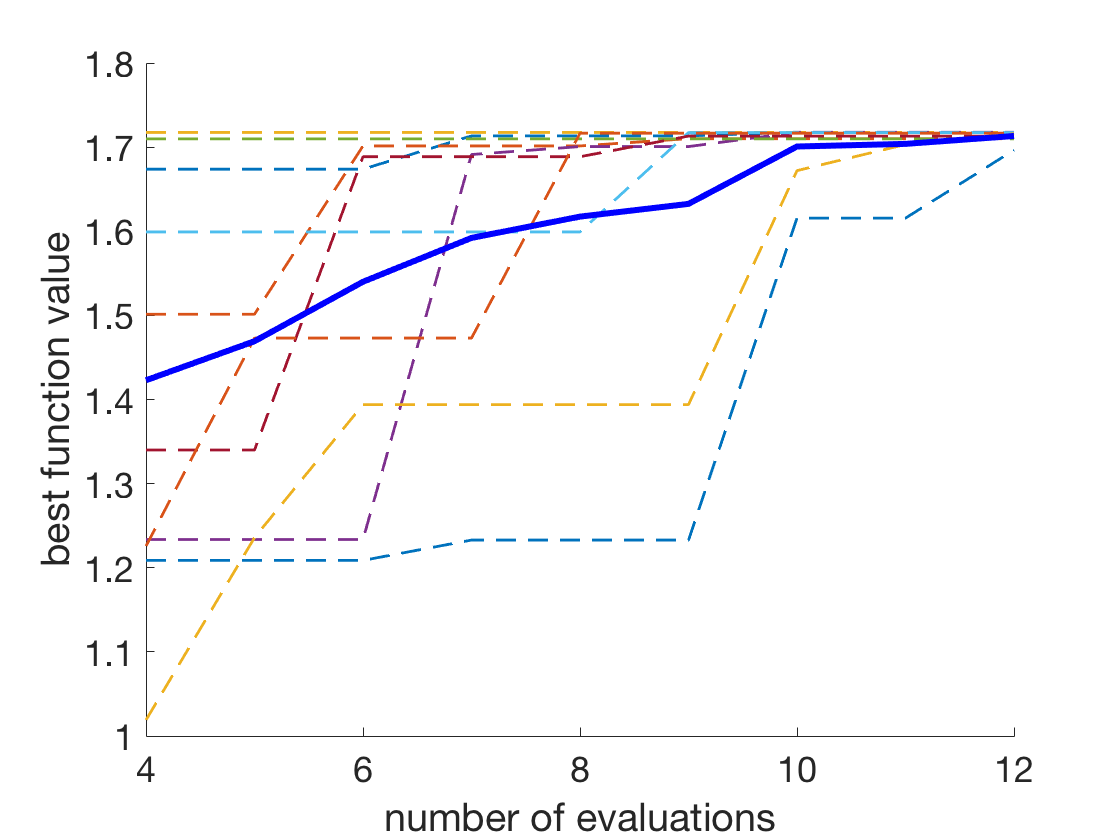}
	\label{fig:1d_monot}
}
\\
\subfloat[Decomposition without monotonicity]
{
	\includegraphics[width=.49\linewidth]{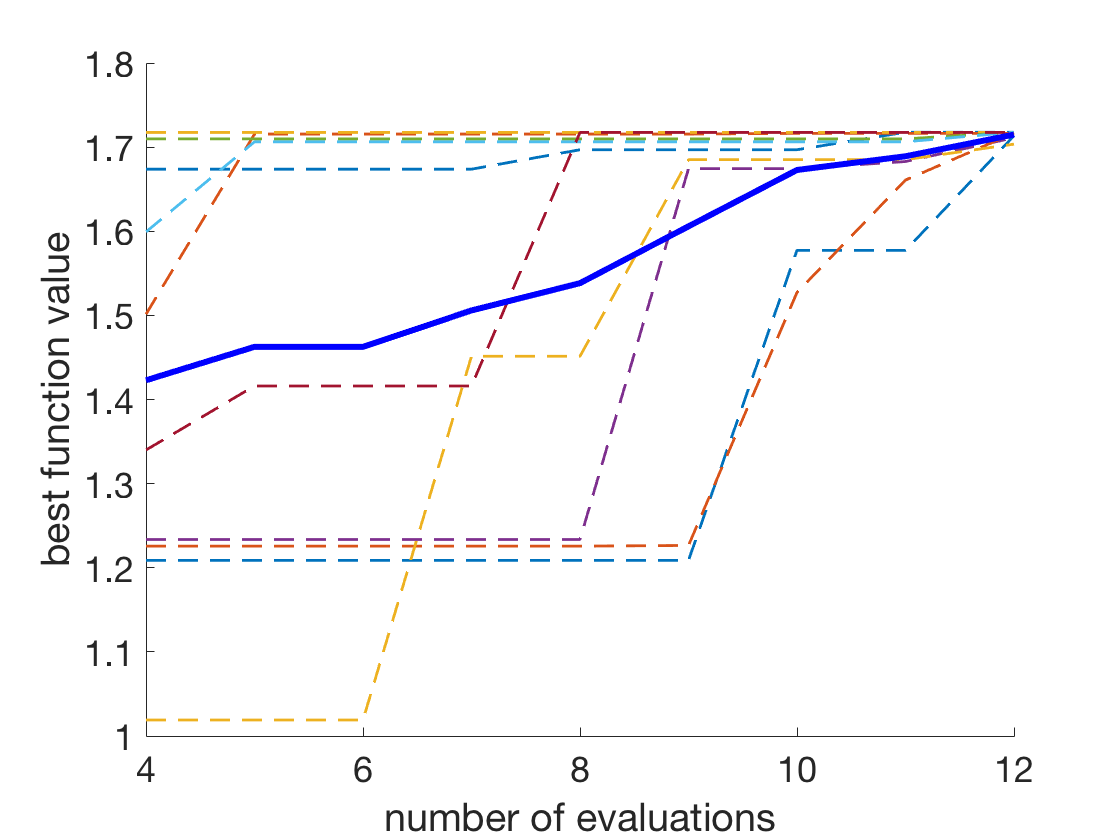}
	\label{fig:1d_non}
}
\caption{Best $f(x)$ value found for the illustrative example versus the number of evaluations for three algorithms: (a) standard Bayesian optimization, (b) decomposition and monotonicity, and (3) decomposition without monotonicity.  An algorithm starts with four evaluations at random points $x \in [0, 1]$ and adds eight further evaluations.  Each dashed line shows one of 10 trials from different random starts; the solid blue line is the average performance of an algorithm.}
\label{fig:1d}
\end{figure}

Figure \ref{fig:1d} shows the result of applying the three algorithms 10 times from four random initial objective function evaluations and adding a further eight evaluations. 
For this problem, we employ a uniform grid of 10 uniform virtual points. We see that modeling decomposed objective functions separately improves the performance. However there is little extra benefit from the monotonicity adaptation here.

\section{Application to Machine Learning Hyperparameter Tuning}
\label{sec:app}
A large class of ML algorithms can be formalized as having three components: (1) a hypothesis set, (2) a loss function (usually highly related to a pre-defined performance measure), and (3) an optimization of the loss function over the hypothesis set with respect to the training set. The ideal utility of a ML algorithm is usually defined as generalization error, which is the expected performance measure of the chosen hypothesis over the underlying distribution. Hyperparameters in ML are often to describe the hypothesis set and loss function.  

The generalization error in a real problem is usually inaccessible. Instead, the validation error is often used as an approximation for the purpose of hyperparameter tuning. Hence the goal of hyperparameter tuning in ML is to optimize the hyperparameters with respect to validation error. Rather than optimizing validation error directly, we propose to optimize the training error plus the difference between validation error and training error. We claim that these component functions are reasonably assumed to be monotonic with respect to ML model complexity and hence the hyperparameters.

For many ML algorithms, the training error (i.e., the performance measure of the chosen hypothesis on the training set) is monotonically decreasing as the hypothesis set increases. The training error is also monotonic with respect to regularizing parameters, which are also part of the loss function, in many cases. Hence  monotonicity of training error with respect to hyperparameters widely exists. 

Analysis of monotonicity of the difference given by validation error minus training error is much harder. To the best of our knowledge, there is no general result. Even monotonicity in expectation does not exist. However the idea that the generalizability (inverse of the difference between generalization and training error) is decreasing with respect to effective model complexity is accepted in the ML community. That view leads to a heuristic argument that assuming monotonicity of validation error minus training error is reasonable.

Consider an ML algorithm $A$ that learns a mapping from $\mathcal{X}$ to $\mathcal{Y}$. Let the training set $X_{tr}$ and $y_{tr}$ be $n$ identical independent (i.i.d.) samples from distribution $D$. Denote the learned mapping as $A(X_{tr}, y_{tr})$. Fix a performance measure $m:\mathcal{Y} \times \mathcal{Y} \to \R$. The training error and generalization error are defined as
\begin{align*}
err &= \frac{1}{n}\sum_i m(A(X_{tr}, y_{tr})(X_{tr,i}), y_{tr,i}), \quad \text{and}\\
e &= \E_{x,y \sim D}[m(A(X_{tr}, y_{tr})(x), y)].
\end{align*}
The generalization error can be rewritten as 
\[
E = \E_{X', y' \sim D^n} \frac{1}{n} \big[ \sum_i m(A(X_{tr}, y_{tr})(X'_i), y'_i)\big].
\]

Notice that the training error and generalization error are of the same form: both involve means for $A(X_{tr}, y_{tr})$, evaluated at datasets $X_{tr}, y_{tr}$ and $X', y'$, respectively. Since the two datasets are i.i.d. samples from the same distribution and $A(X_{tr}, y_{tr})$ is optimized for data set $X_{tr}, y_{tr}$, the training error should be more sensitive to the model complexity compared to thegeneralization error. Therefore monotonicity of the difference between generalization error and training error is plausible.

In manual ML hyperparameter tuning, experts implicitly use the same approach. They tune the hyperparameters using monotonicity assumptions and relativities between target error, training error, and validation error minus training error  \citep[pp.\ 424, 425]{goodfellow2016deep}.

\subsection{Elastic Net Regularized Linear Regression}
Consider the elastic net regularized linear regression algorithm. Let $X_{tr}$ and $y_{tr}$ be training sets. The elastic net regularized linear regression tries to find a linear predictor with sparse coefficients; nonzero coefficients are penalized to be smaller in magnitude. It optimizes over all linear functionals a combination of goodness of fit, the 2-norm of the coefficients, and the 1-norm of the coefficients. Specifically, it finds $\hat{\beta}$ to minimize the loss
\begin{align*}
l(\beta) = \frac{1}{2|y_{tr}|} \|y_{tr} - \beta X_{tr}\|_2^2 + \lambda P_\alpha(\beta), \label{loss}
\end{align*}
where $P_\alpha(\beta) = \frac{1-\alpha}{2}\|\beta\|^2_2+\alpha\|\beta\|_1$. We adopt the common performance measure mean square error scaled by $\frac{1}{2}$ for consistency
\begin{align*}
\mse(\beta, X, y) = \frac{1}{2|y|} \|y - \beta X\|_2^2.
\end{align*}
Note that the loss function $l$ consists of two parts: the first is the mean square error on the training set, which measures the goodness of fit, while the second,  $\lambda P_\alpha(\beta)$, is a regularization term.  In the regularization the 1-norm forces sparsity of $\beta$, and the 2-norm constrains the magnitude of $\beta$. The regularization term can be considered as representing prior knowledge of the underlining function and can prevent over fitting for many cases. For more details of elastic net regression, one may refer to \citet{zou2005regularization}.

Given the validation sets $X_v$ and $y_v$, the goal of hyperparameter tuning for this problem is to minimize $\mse(\hat{\beta}, X_v, y_v)$ with respect to $\alpha$ and $\lambda$, where $\hat{\beta} = \argmin_{\beta} l(\beta)$.

\begin{figure}[h]
\centering
\subfloat[Training error]
{
	\includegraphics[width=.50\linewidth]{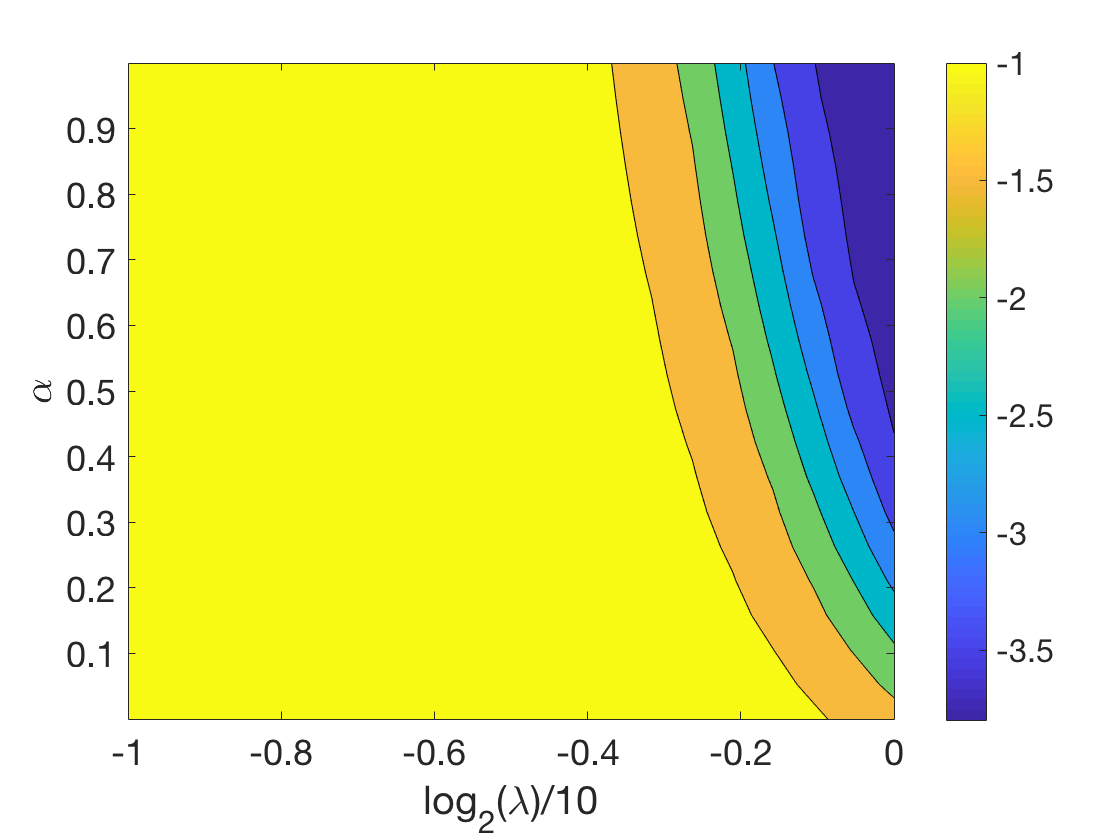}
	\label{fig:EN_tr}
}
\subfloat[Validation error - training error]
{
	\includegraphics[width=.50\linewidth]{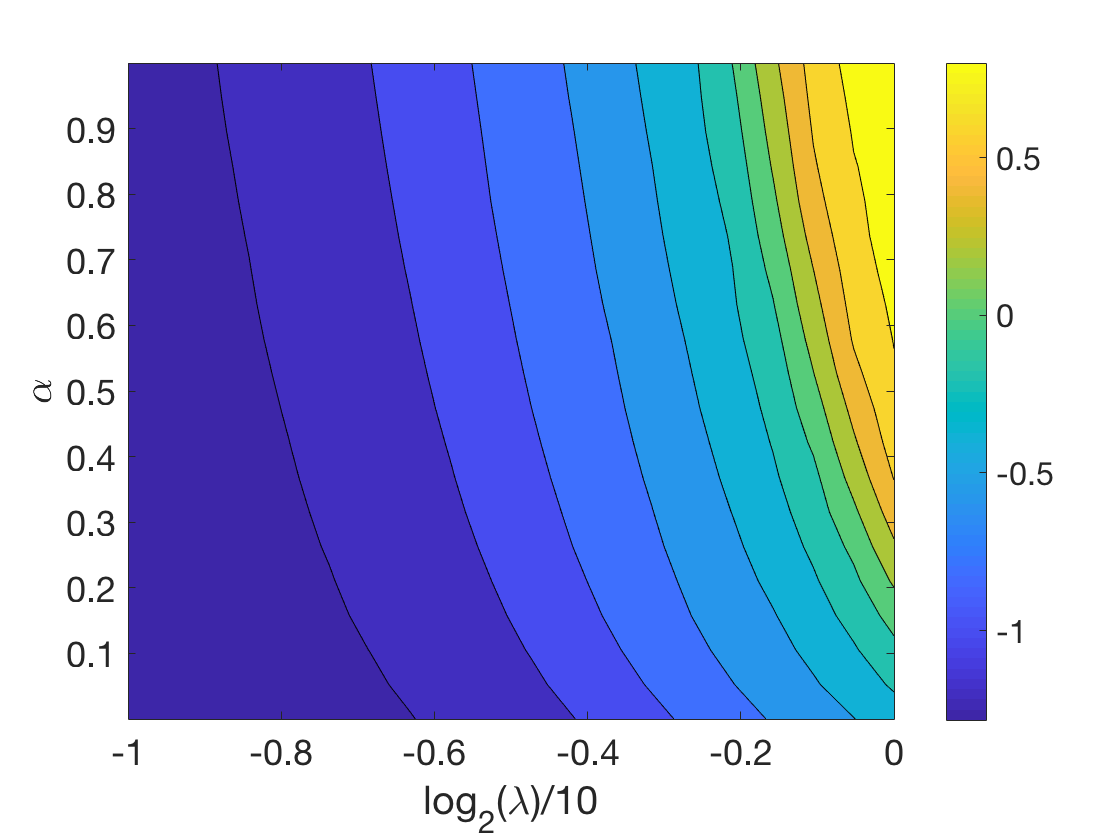}
	\label{fig:EN_de}
}
\\
\subfloat[Validation error]
{
	\includegraphics[width=.5\linewidth]{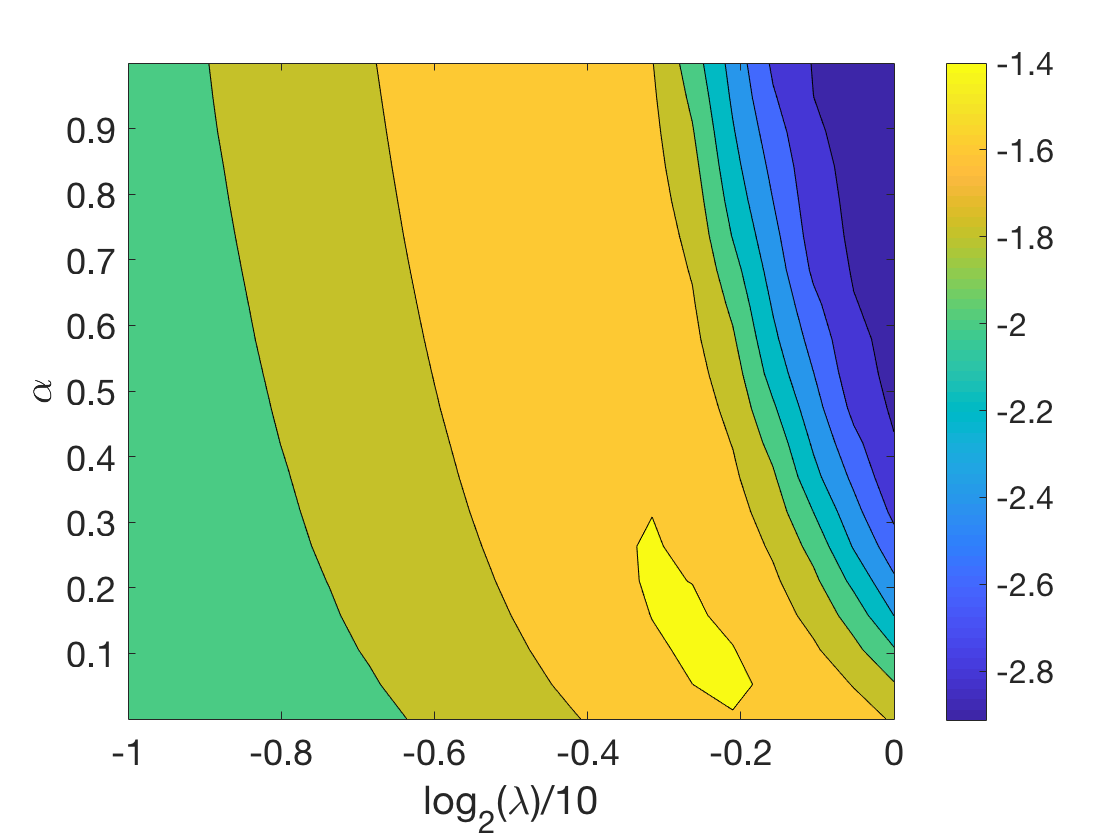}
	\label{fig:EN_val}
}

\caption{Validation error of the elastic net linear regression problem as a function of the hyperparameters $\alpha$ and $\lambda$, and its decomposition. (a) and (b) show the training error and validation error minus training error, respectively, which are assumed to be monotonic. Since we describe Bayesian optimization for maximization and validation error is to be minimized, negative error values are shown. The yellow region is desired for this hyperparameter turning problem.}
      \label{fig:EN_errors}

\end{figure}

In the following experiment, all elements of $X_{tr}$ and $X_v$ are sampled independently from the standard normal distribution, with 200 examples in each set and 100 features for each example. We define the underlining linear transformation by a $100 \times 1$ vector $C$ with 50 zero elements; the remaining 50 nonzero elements are drawn from a normal distribution with mean zero and standard deviation 0.22. Data $y_{tr}$ and $y_v$ follow independent normal distributions with means $C X_{tr}$ and $C X_v$, respectively, and standard deviation 1. The parameters for generating the data are chosen to make the optimization problem nontrivial. The hyperparameter ranges of interest are $\alpha \in [0,1]$, and $\lambda \in 2^{([-10,0])}$. Figure \ref{fig:EN_val} shows the validation error as a function of $\alpha$ and $\lambda$.
Since we describe Bayesian optimization for maximization and validation error is to be minimized, negative error is shown.

As we described in the general case, to optimize the validation error, we decompose it as the sum of the training error and the difference between validation error and training error. From Figures \ref{fig:EN_tr} and \ref{fig:EN_de} we see that the two decomposed functions are (at least roughly) monotonic, which satisfies our assumptions. 

\begin{figure}[h]
\centering
\subfloat[Standard]
{
	\includegraphics[width=.5\linewidth]{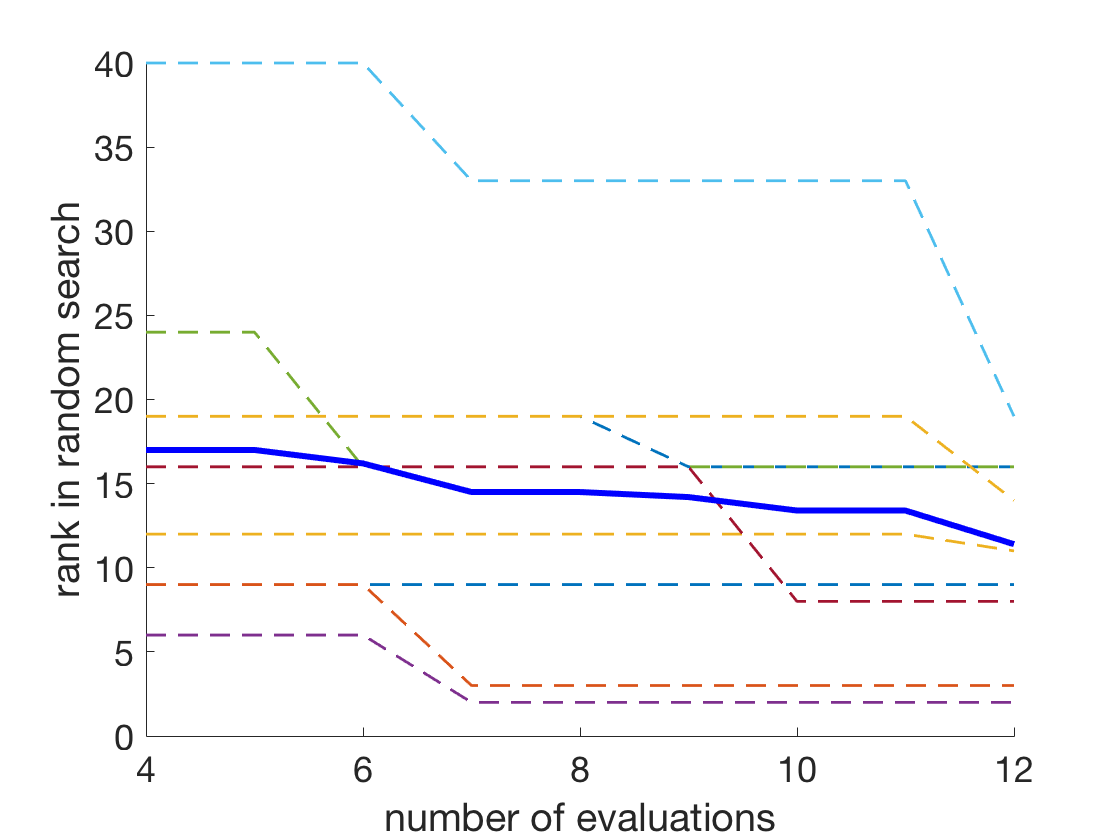}
	\label{fig:EN1}
}
\subfloat[Decomposition and monotonicity]
{
	\includegraphics[width=.5\linewidth]{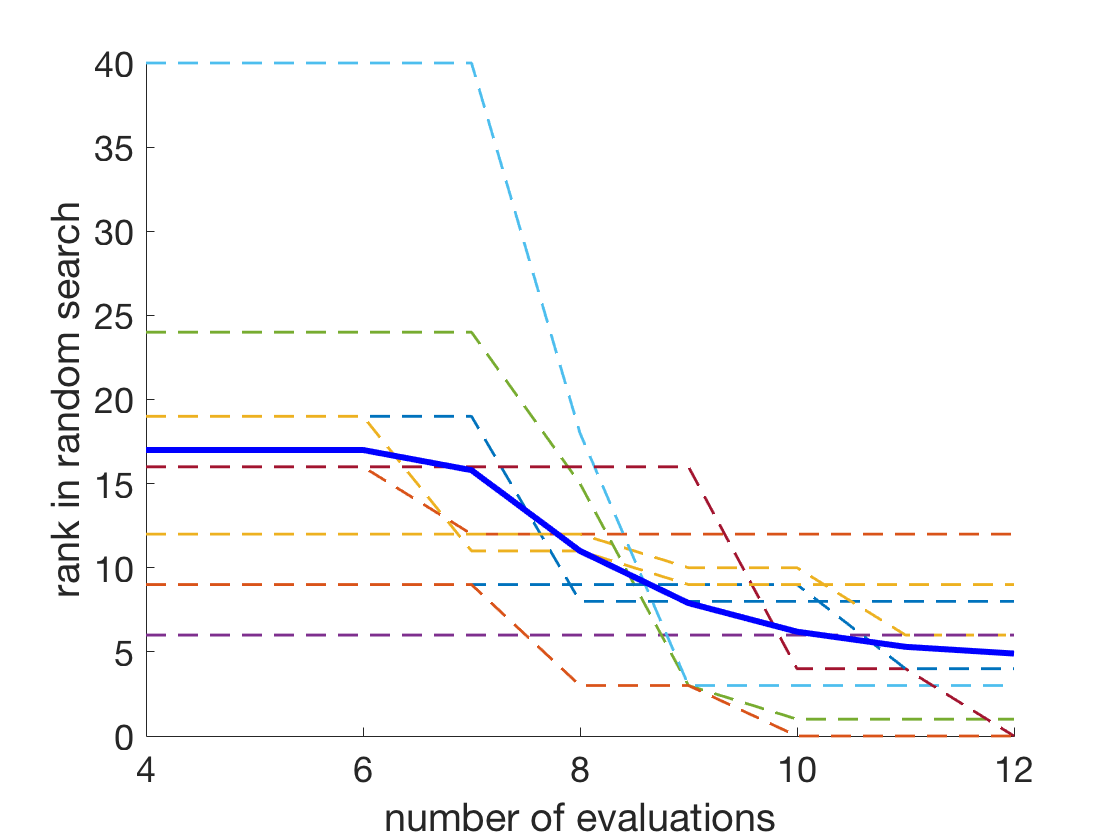}
	\label{fig:EN2}
}
\\
\subfloat[Decomposition without monotonicity]
{
	\includegraphics[width=.5\linewidth]{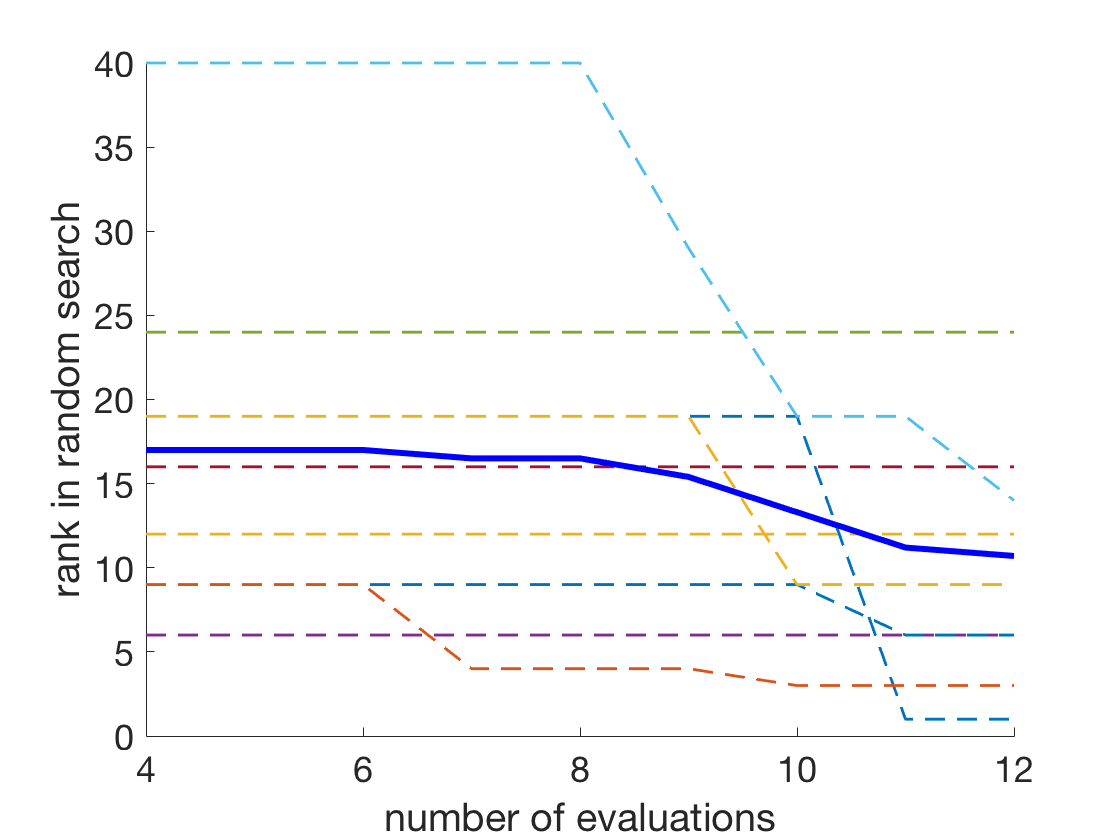}
	\label{fig:EN3}
}
\subfloat[Mean squares]
{
	\includegraphics[width=.5\linewidth]{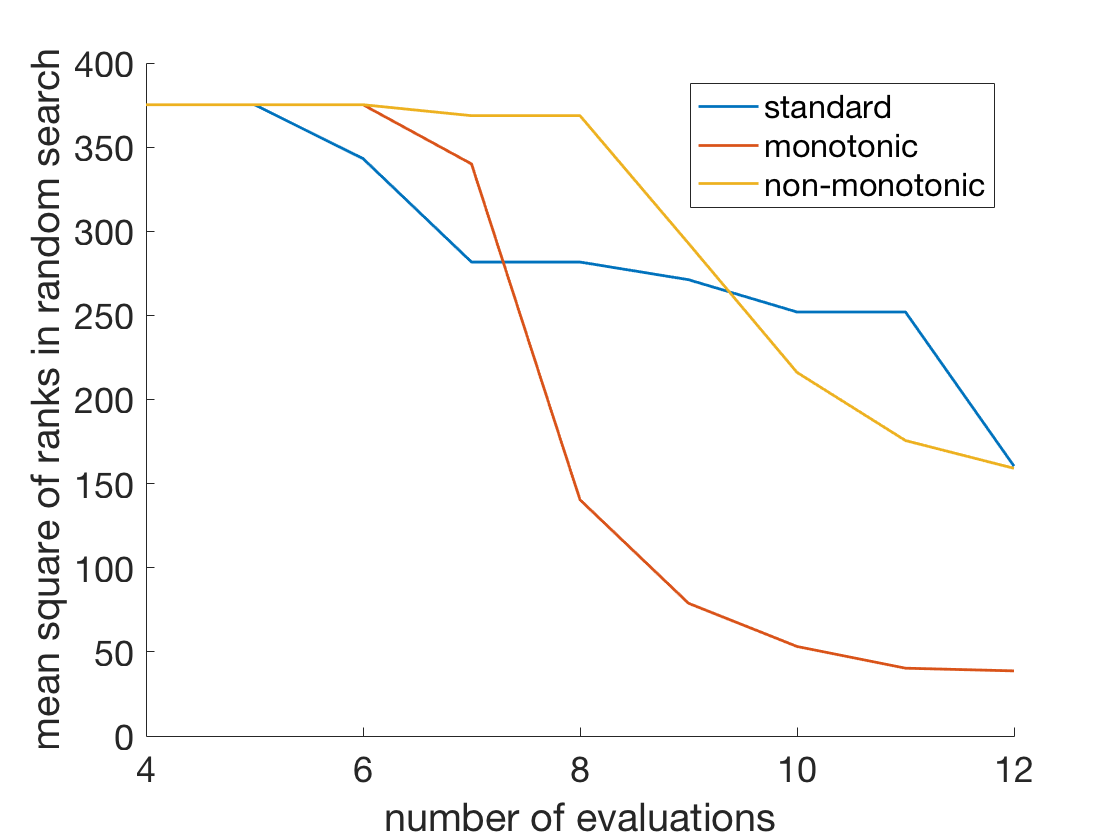}
    \label{fig:ENSM}
}
\caption{Experimental result of the three algorithms applied on the elastic net hyperparameter tuning problem with 10 trials. Performance is evaluated as ranking of objective function value in 100 random search. In (a), (b) and (c), x-axis and y-axis are corresponding to number of objective function evaluations and best ranking in random search respectively. Each dash is corresponding to one randomly initialized trial. The solid line is average over 10 trials. (d) is the mean square of the rankings over 10 trials.}
        \label{fig:EN_result}
\end{figure}

Again, we compare three algorithms: standard, decomposition and monotonicity, and decomposition without monotonicity.  Each is  applied 10 times, starting from different initial function evaluations at four random points. A $10 \times 10$ grid of virtual points is used in monotonicity modeling. From the experiment we observe that the standard and non-monotonic algorithms start by  evaluating the functions at the corners, whereas the monotonicity modeling algorithm does not always do so. This is reasonable since the monotonicity modeling algorithm has more prior knowledge of the objective function. 

Figure \ref{fig:EN_result} shows the results from the three algorithms for the 10 trials. Since the objective function value is hard to distinguish near optimal values, we measure the performance of an algorithm as its rank among 100  evaluations at random points.  
Figure \ref{fig:EN1}, \ref{fig:EN2} and \ref{fig:EN3} show the best ranking at each iteration for the three algorithms. 
We see that the performance of the algorithm employing decomposition and monotonicity dominates the other two statistically. For many trials of the standard and decomposition-only algorithms, 12 objective function evaluations do  not lead to significant improvement. This indicates that these algorithms require more evaluations to learn an accurate model of the objective function. In contrast, the algorithm exploiting  decomposition and monotonicity makes significant improvement in most of the trials.  Figure \ref{fig:ENSM} plots the mean square of the rankings in \ref{fig:EN1}, \ref{fig:EN2} and \ref{fig:EN3} over 10 trials at each iteration. The mean square of a set of real valued samples equals the sum of the squared mean and the variance of the samples. Therefore a smaller mean square value implies a smaller mean on average, and/or less variation in rankings, which is what we desire for optimization. 

\subsection{Random Forest on Adult Dataset}
In this subsection, we apply and compare the three algorithms when tuning random forest hyperparameters for the Adult dataset. The Adult dataset contains 32561 training examples. For each instance, there are 14 features describing a person and one binary variable that indicates a salary greater than 50K for classification \citep{Lichman:2013}. Random forest classification has been successfully applied to this dataset \citep{fan2005effective}. 

Random forest classification is an ensemble method that operates by constructing a multitude of decision trees at training time.  To classify a new instance, it outputs the class that is the mode of the classes predicted by the individual decision trees (majority vote). To balance the goodness of fit and generalizability of each decision tree, we control the model complexity by tuning  three hyperparameters: maximum split nodes, minimum leaf size, and number of features to use at each split. In this experiment, we adapt the \textbf{TreeBagger} implementation of random forest in the \citet{MatlabOTB}. 
We generate random forests with 10 decision trees and set other tuning parameters at default values. The training sets $X_{tr}, y_{tr}$ and validation sets $X_{val}, y_{val}$ comprise $70\%$ and $30\%$ of the dataset, respectively. Classification error is used for hypothesis performance measurement. The hyperparameter optimization problem is to find the best $\theta_1, \theta_2, \theta_3 \in [0,1]$ such that
\begin{align*}
&\error(h, X_{val}, y_{val}) \\
= & \frac{1}{|X_{val}|}\sum_i |h(x_i) - y_i| \\
= & \error(h, X_{tr}, y_{tr}) \\
&+ (\error(h, X_{val}, y_{val}) - \error(h, X_{tr}, y_{tr}))
\end{align*}
is minimized, where $h$ is the random forest generated with maximum split nodes equals $2^{\theta_1 \log_2(|X_{tr}|)}$, minimum leaf size equals $2^{\theta_2  \log_2(|X_{tr}|)}$ and  number of features to use at each split equals $2^{\theta_3\log_2(14)}$. The integer hyperparameters here and in the next experiment are rounded to  the nearest integers.

\begin{figure}[h]
\centering
\subfloat[Standard]
{
	\includegraphics[width=.5\linewidth]{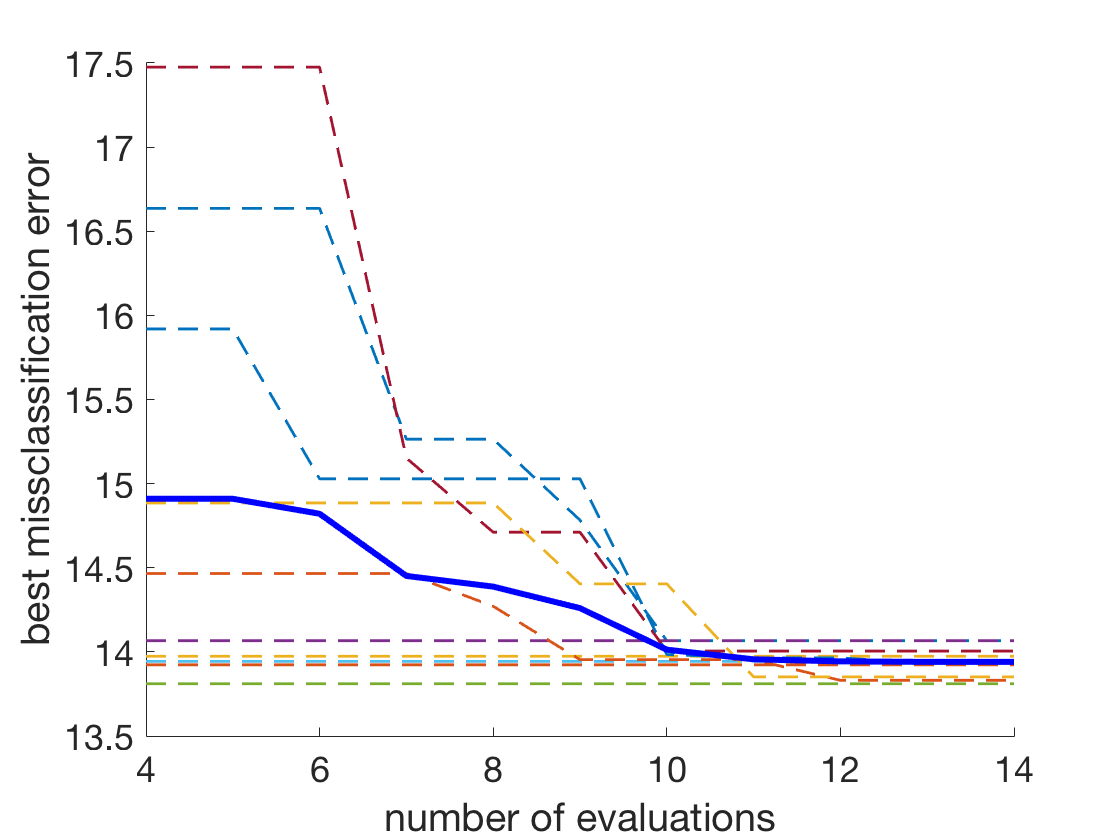}
	\label{fig:RF1}
}
\subfloat[Decomposition and monotonicity]
{
	\includegraphics[width=.5\linewidth]{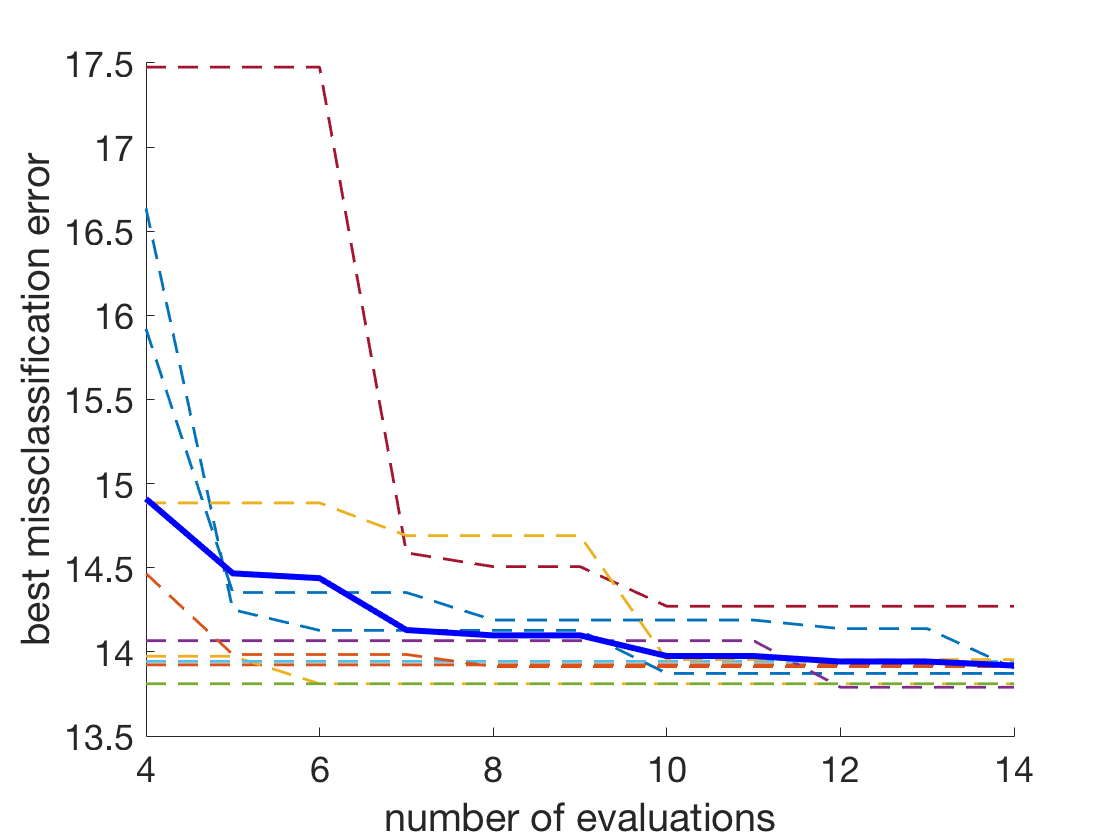}
	\label{fig:RF2}
}
\\
\subfloat[Decomposition without monotonicity]
{
	\includegraphics[width=.5\linewidth]{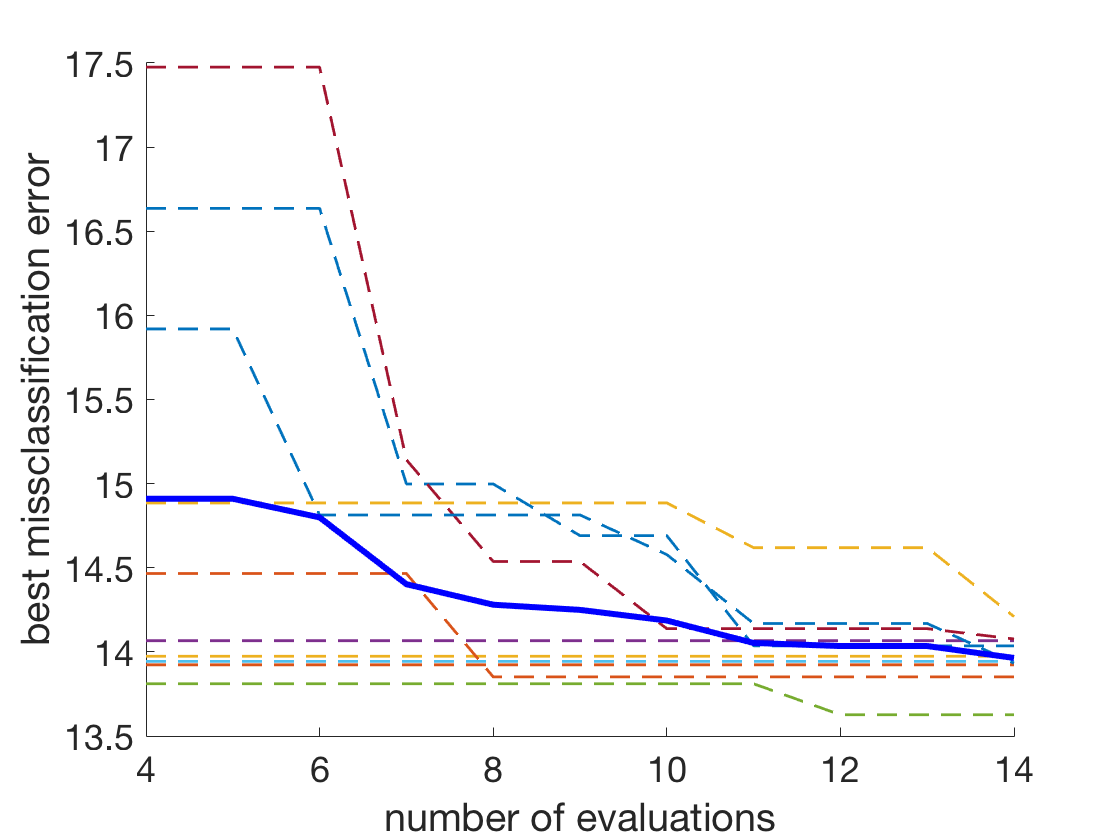}
	\label{fig:RF3}
}
\subfloat[Mean squares]
{
	\includegraphics[width=.5\linewidth]{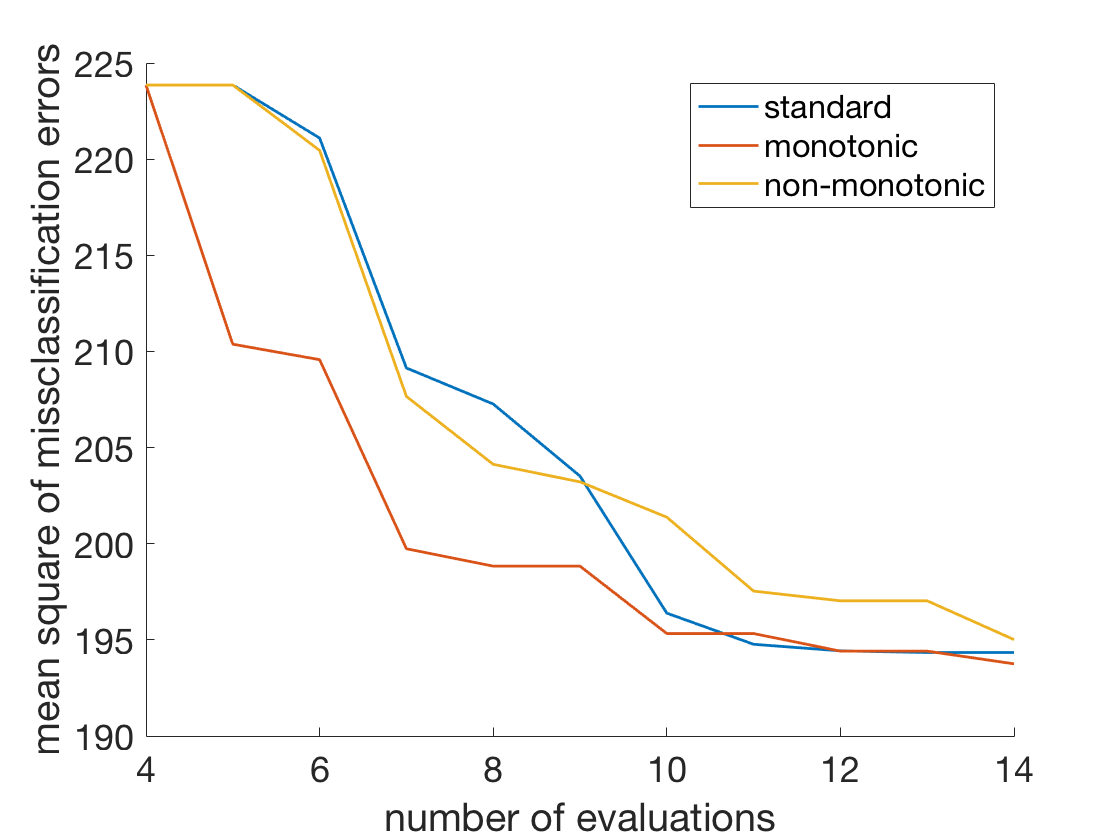}
    \label{fig:RFMS}
}
\caption{(a), (b), and (c) show the best misclassification errors versus number of evaluations for three algorithms applied to the random forest hyperparameter tuning problem. Each dashed line corresponds to one of 10 randomly initialized trials. The solid line is the average over the 10 trials. (d) is the mean square of the objective function values over the 10 trials.}
        \label{fig:RF_result}
\end{figure}

We apply the three algorithms to this problem. Virtual points in our algorithm are on a $4 \times 4 \times 4$ grid. Figure \ref{fig:RF_result} shows experimental result of running the three algorithms 10 times with four initial objective function evaluations. The performance of result of distributions at termination are not clearly distinguishable. However by looking at both the validation error and the mean square measure we see that our algorithm improves faster, and always dominates other methods.

\subsection{Convolutional Neural Network on CIFAR-10 Dataset}
Deep learning has enjoyed wide success in many areas in the past decade. One of the challenging problems to apply deep learning algorithms is to select the relatively large number of hyperparameters. Some of them define the effective model complexity, and it is reasonable to assume monotonicity as discussed in general. In this work, we tune three hyperparameters: the $L_1$ regularizer $\alpha_1 \in [0,1]$, the $L_2$ regularizer $\alpha_2 \in [0,1]$, and an integer $n \in [4,68]$ that controls the number of neurons for a 5-layered convolutional neural network (CNN).  The target problem is the CIFAR-10 image classification dataset \citep{krizhevsky2009learning}. A systemic study of CNN image classification and its application to this data set can be found in \citet{krizhevsky2012imagenet}. 

We construct CNNs with one input layer, two convolutional layers, a fully connected layer, and a softmax output layer. The two convolutional layers consist of $n$ and $2n$ feature maps with 3 by 3 kernel size, ReLU activation functions, and 2 by 2 max pooling. The  fully connected layer adapts ReLU activation functions with $n^2/2$ neurons. Each CNN is trained by stochastic gradient decent on $L_1$ and $L_2$ regularized cross entropy with constant learning rate $0.02$, decay factor  $10^{-6}$, momentum $0.9$ and number of epochs equals to 30. Since training the whole data set takes substantial time, for demonstration purposes we randomly select 2000 examples as a training set. We use cross entropy as the performer measure on the validation set.

\begin{figure}[h]
\centering
\subfloat[Standard]
{
	\includegraphics[width=.475\linewidth]{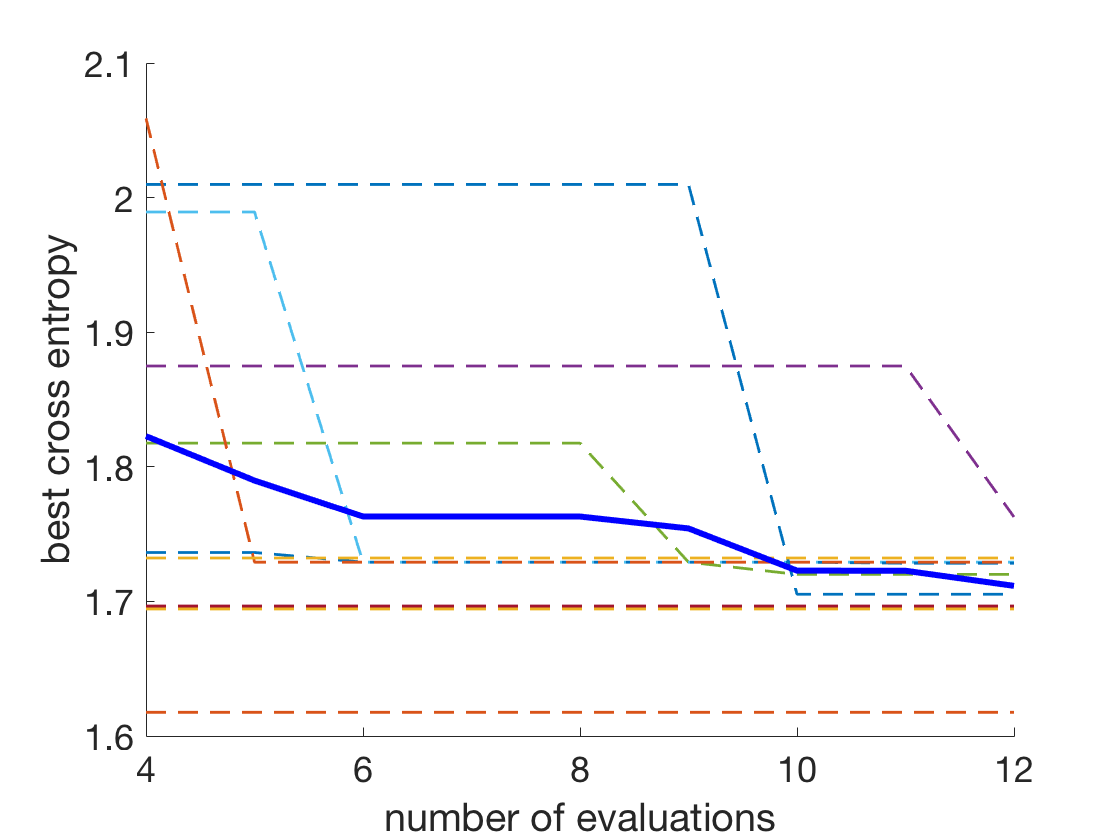}
	\label{fig:cnn1}
}
\subfloat[Decomposition and monotonicity]
{
	\includegraphics[width=.475\linewidth]{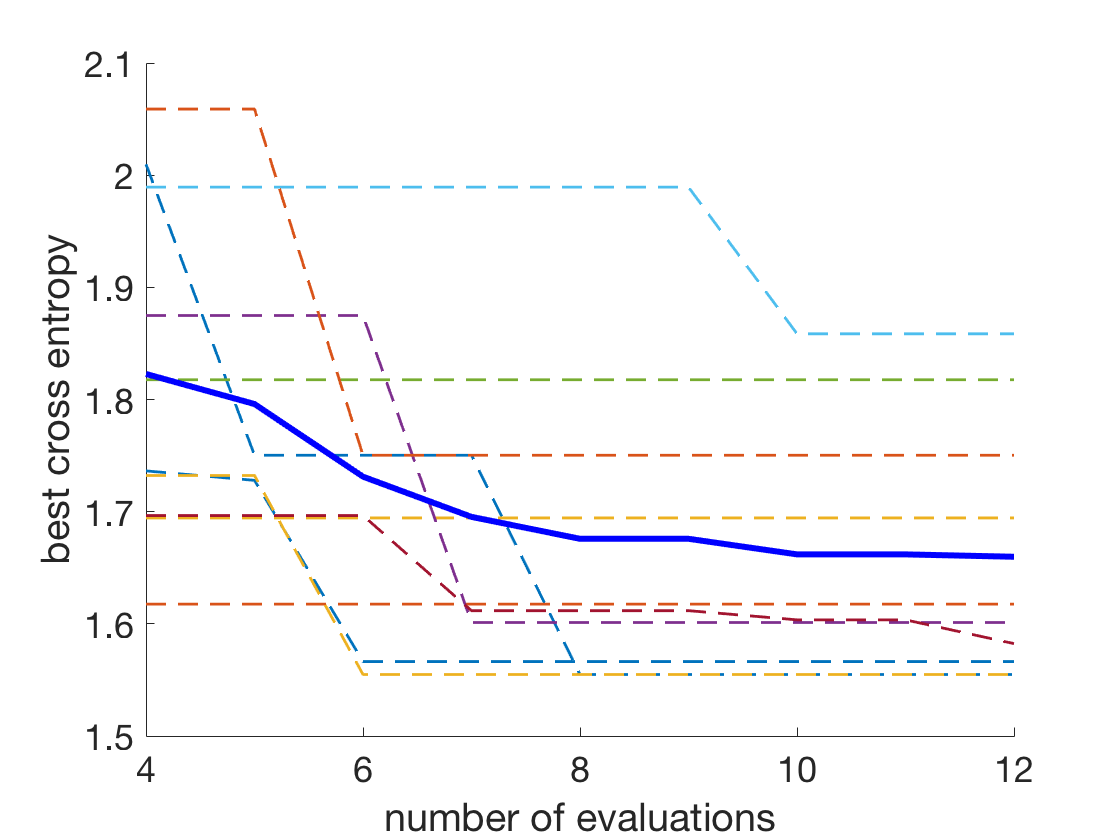}
	\label{fig:cnn2}
}
\\
\subfloat[Decomposition without monotonicity]
{
	\includegraphics[width=.475\linewidth]{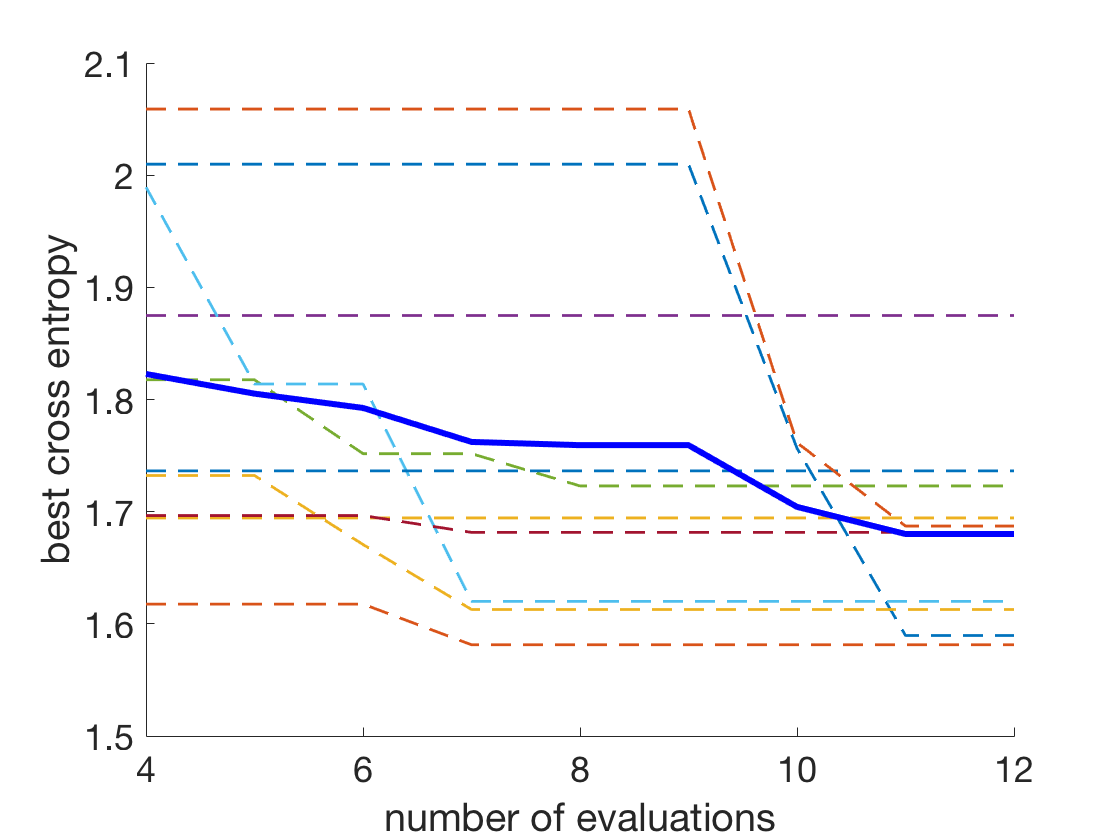}
	\label{fig:cnn3}
}
\subfloat[Mean squares]
{
	\includegraphics[width=.475\linewidth]{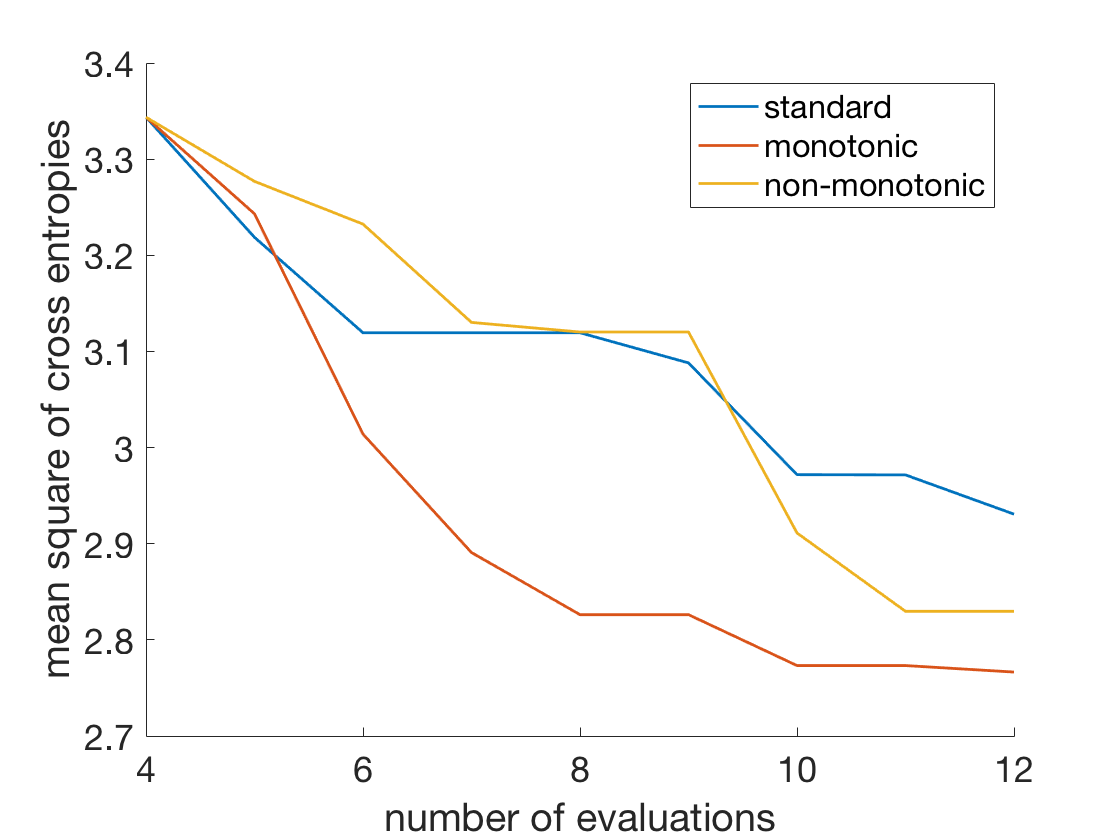}
    \label{fig:cnnMS}
}
\caption{(a), (b), and (c) show the best evaluated cross entropy value versus number of evaluations for three algorithms applied to the convolutional neural network hyperparameter tuning problem. Each dashed line corresponds to one of 10 randomly initialized trials. The solid line is the average over the 10 trials. (d) is the mean square of the objective function values over 10 trials.}
        \label{fig:cnn_result}
\end{figure}

Figure \ref{fig:cnn_result} shows the results from applying the three algorithms to the CNN problem. Optimization settings are identical to those for the random forest problem.   We see that the performance of the algorithm imposing monotonicity dominates the other two statistically. All trials of the standard algorithm fail to achieve an objective function below 1.7. Decomposing the objective function improves performance,
and also encoding monotonicity information gives the highest rate of objective values below 1.7. We also observe that the algorithm exploiting decomposition and monotonicity in modeling makes improvement earlier compared to the other two algorithms. 

\section{Discussion and Future Works}
\label{sec:last}
This work is inspired by manual strategies for hyperparameter  tuning of an ML algorithm. 
That strategy adjusts the hyperparameters for more model complexity when observing training error greater than target error, or reduces complexity if training error is reasonably small but validation error is not satisfied. One can do this because of prior knowledge that training error and the generalizability are both decreasing with respect to algorithm model complexity.  The strategy implicitly decomposes the validation error into training error plus training error minus validation error (generalizability is the inverse of this difference). Our algorithm modifies the modeling part of Bayesian optimization to exploit this prior knowledge and the decomposition. The hope is that the decomposition of validation error simplifies the tuning problem by creating two functions that are easier to model, i.e., less nonlinear for example.  Moreover, approximate monotonicity provides further useful information for modeling and optimization. The experiments indicate that for ML hyperparameter tuning monotonicity information provides more improvement.

There exist two approaches for monotonic GP regression. \citet{riihimaki2010gaussian} and \citet{wang2016estimating} enforced monotonicity by inserting virtual observations of derivative processes, and \citet{lin2014bayesian}  projected  GPs onto a constrained space. An important question  is where to insert the virtual points. The trivial method that we employed using a grid results in having exponentially many virtual points with respect the objective function's domain dimension.  Training a GP requires $O(n^3)$ computation with a large constant, however, where $n$ is the number of observations. With limited computational power, methods that can assign virtual points as requested to fill the space more sparsely is more preferred, for example, Latin hypercube sampling \citep{McKBecCon1979}. A domain-specific adaptive method was proposed by \citet{wang2016estimating}. It sequentially determines one virtual point at a time at the location where the derivative process has the lowest (or highest) value. However this method itself requires repeating the training process and optimization of the learned process a linear number of times with respect to requested number of virtual points.  Thus, it is more computationally expensive that non-adaptive methods.

\citet{lin2014bayesian}'s method first learns the posterior process without monotonicity information. Then it samples functions from the posterior process, and projects them into the monotonicity constrained space. The projection is defined to minimize an integral of square difference. It is not clear how samples from the posterior process are represented, but we believe a (possibly approximate) closed form projection can be derived with compact sample representation. If a closed formed solution is available, the monotonic GP regression required by our algorithm should share the same order of computation as standard GP regression.

For ML hyperparameter tuning problems, since the monotonicity of validation error minus training error is not guaranteed even in expectation, one may wonder if imposing a monotonicity constraint is appropriate. Here we make an argument that for an objective function that decompose to two functions with one of them is known to be monotonic, there is an extra reason to model the other function as monotonic beside the behavior of the function itself. Denote the decomposed functions by  $f_1$ and $f_2$, with $f_1$  known to be monotonic. Consider the simple case where the domain of the objective function has dimension one. Without loss of generality assume $f_1$ to be monotonically increasing. Then for the interval $[a, b]$ where $f_2$ is increasing, $f_1(x) + f_2(x) \leq f_1(b)+f_2(b)$ for all $x \in [a,b]$. Hence a region where $f_2$ is not monotonically decreasing is not of interest. However, since we do not know in what region $f_2$ is increasing, enforcing  $f_2$ to be monotonic may have a negative effect on accuracy of prediction when $f_2$ is in its decreasing region. This argument can be easily generalized to higher dimensions.

Although this work is motivated for and focused on machine learning hyperparameter problems, our algorithm is designed for any optimization problem with an objective function that can be decomposed as a sum of functions with monotonicity constraints. An example is to optimize a product's price for maximum gain.  A simple but widely used model for this problem is that gain equals $d \times (p - c)$ with $d$ for demand, $p$ for price and $c$ for cost. Cost is assumed to be constant, demand is a monotonically decreasing function of price. Hence optimize $d \times (p-c)$ is equivalent to optimize $\log(d) + \log(p-c)$ which are monotonic with respect to $p$. Evaluating this is sometimes expensive since $d$ given $p$ may be modeled using complicated game theory models. A more complex example was given by \citet{goettler2011does}, where they proved three monotonicities of decomposed expected gain given the current product price and amount of investment for CPU companies.

In conclusion, we propose a Bayesian optimization algorithm for objective functions that can be decomposed as a sum of functions with monotonicity constraints. We demonstrate that many machine learning hyperparameter tuning problems fall into this family. We provide experimental results  for an artificially designed problem, regularized linear regression, random forest classification, and a convolutional neural network for image classification.  They show our algorithm outperforms the standard Bayesian optimization method based on direct modeling of the validation error. 
\bibliography{references}
\bibliographystyle{icml2018}

\end{document}